# Multi-agent Interactive Prediction under Challenging Driving Scenarios

Weihao Xuan
School of Mechanical Engineering
University of Leeds
Leeds, UK
email: mn16whx@leeds.ac.uk

Ruijie Ren
School of Mechanical Engineering
University of Leeds
Leeds, UK
email: cn16rr@leeds.ac.uk

*Abstract*—In order to drive safely on the road, autonomous vehicles are expected to predict future outcomes of its surrounding environment and react properly. In fact, many researchers have been focused on solving behavioral prediction problems for autonomous vehicles. However, very few of them consider multi-agent prediction under challenging driving scenarios such as urban environments. In this paper, we proposed a prediction method that is able to predict various complicated driving scenarios where heterogeneous road entities, signal lights, and static map information are taken into account. Moreover, the proposed multi-agent interactive prediction (MAIP) system is capable of simultaneously predicting any number of road entities while considering their mutual interactions. A case study of a simulated challenging urban intersection scenario is provided to demonstrate the performance and capability of the proposed prediction system.

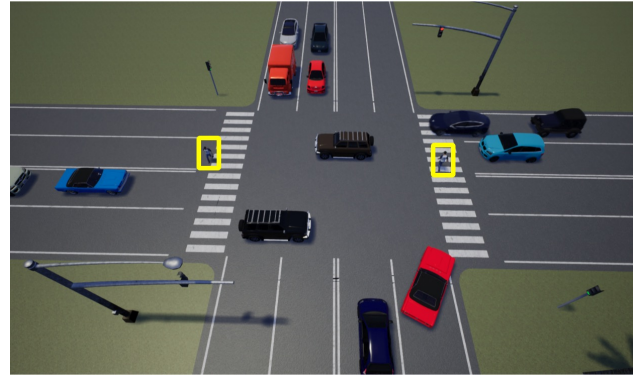

Figure 1. Illustration of an urban intersection driving scenario. Pedestrians are bounded by the yellow boxes.

## I. INTRODUCTION

### A. Motivation

For autonomous vehicles, the ability to drive safely in a sophisticated driving environment is required. Since the road entities and traffic lights are dynamic in the driving environment, autonomous vehicles should be able to examine and react to these situations while ensuring their own safety. In fact, Advanced Driver Assistance System (ADAS) has been designed to assist human drivers and enhance the driving experience. In recent decades, approaches to predict future behaviors of on-road vehicles have made big progress, especially under simple scenarios such as highways. However, for complex driving environments such as urban areas in Fig. 1, the challenges for the researchers to design comprehensive prediction algorithms still remain. In this work, we would like to build a prediction structure that can be easily adapted to any complicated urban driving scenarios with given semantic map information. In particular, we aim at forecasting future behaviors of all types of road entities while considering their mutual interactions.

### B. Related Works

In the field of computer vision and robotics, prediction of future behaviors for vehicles has been studied by different approaches. For example, to predict future behaviors and states for vehicles, traditional methods such as dynamic, kinematic model and Intelligent Driver Model (IDM) have been designed [1]-[3]. However, these methods have big limitations that the predictions are made by estimating each step recursively under underlying assumptions such as constant velocity and constant acceleration. In order to make a better prediction for the model with uncertainty, more frameworks are developed, such as Gaussian Processes (GP), Hidden Markov Models (HMM) [4] and Monte Carlo sampling [5] which is designed to consider various motion patterns. These approaches have good results in motion prediction for single vehicle driving scenarios without consideration of interactions with other vehicles. However, these methods are unreasonable since behaviors of the predicted vehicle are influenced by surroundings in the real world.

In order to take interactions into account, there are works extracting features from the environment which have a potential effect on the ego vehicle. Most of these works use learning-based methods to consider interactions among vehicles and other entities [6]-[8]. For example, [7] predicts the most likely future longitudinal and lateral trajectories for the ego vehicle on a highway surrounded by the other nine vehicles using LSTM. Authors in [8] combine an LSTM-based structure with occupancy grid maps to predict the behavior of vehicles taking the lane change behaviors of surrounding vehicles into account. Works such as [9] and [10] also consider the lane change interaction among multiple vehicles. However, these works only focus on simple scenarios such as highways where the moving direction of all the vehicles is the same, and the interactions between vehicles are straightforward since the speed of vehicles on highways is bounded and does not have a significant variation. Approaches from these works cannot be adapted to an urban environment where driving conditions and interactions are more complicated.

Several works have been done to establish more robust predicting systems under a complicated driving context. Approaches from [11] use a stochastic filtering framework to predict the behaviors and trajectories for vehicles in a merged intersection. [12] and [13] build stereo-vision systems to track and predict future trajectories of vehicles driving in roundabouts scenarios. [14] considers an urban area with 4-way intersections by using HMM, and [15] uses Monte Carlo method to predict multimodal trajectories in urban intersections. In these works, interactions between vehicles in complex scenarios are studied. However, in the real urban driving situations, vehicles are also greatly influenced by dynamic traffic signals and pedestrians passing through the streets which are not considered by these works.

Indeed, interactions between vehicles and heterogeneous road entities in intersections were considered using vehicle-to-vehicle (V2V) communication [16] where all the entities are assumed connected by wireless waves etc. However, these connections between road entities were not assumed in this work which makes our approach more suitable for the current real-world situation where no preset connections are available.

In this work, we consider interactions between the predicted vehicle and heterogeneous road entities that might have potential influences on the predicted vehicle. In our approach, we aim to build an autonomous system that could predict future behaviors for all vehicles under a challenging driving scenario. In fact, we claim that for any driving scenario, it can be categorized into two parts: *static information* and *dynamic information*. For a single vehicle driving on the street, it should obey traffic rules by considering static information such as lane directions, crosswalk, sidewalk, and infeasible driving regions. When the vehicle is surrounded by other moving entities, dynamic information including traffic lights and states of other road entities such as vehicles, pedestrians or cyclists, should be taken into account simultaneously. By incorporating all this information, we are able to design a generic prediction architecture that can be easily adapted to various challenging driving environments.

*C. Contribution*

In this paper, a Multi-agent Interactive Prediction (MAIP) system is proposed for autonomous vehicles to predict future behaviors of every road entity while taking interactions into account. We utilize a complicated urban intersection scenario to implement and examine our proposed system. The main contributions of this work can be summarized in four folds:

- Taking into account various environment information such as heterogeneous road entities, lane direction, and traffic light, where the number of heterogeneous entities in the prediction system is not limited.

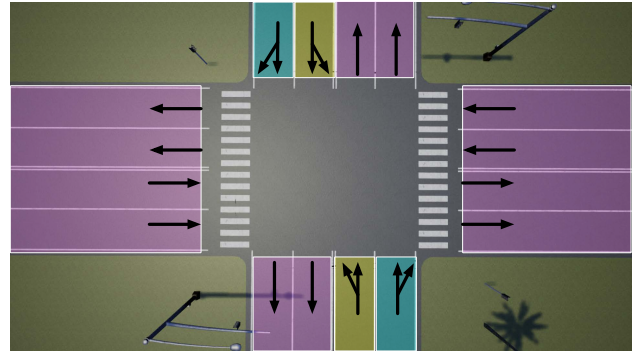

Figure 2. The top-down view of intersection area where *purple* denotes $L_1$, *yellow* and *blue* represent $L_2$ and $L_3$ respectively.

- Introducing an adaptive prediction system by utilizing grid maps to encode static and dynamic environments.
- Considering multimodal properties of predicted distributions.
- Leveraging the masking mechanism to improve learning efficiency.

II. PROBLEM FORMULATION

In this problem, our goal is to solve trajectory prediction problems for vehicles considering environment interactions in an urban intersection area. To increase the environment complexity, we consider three types of driving lanes $L$ for the intersection: $\{L_i : \{F, L, R\}, \ i = 1, 2, 3\}$. Here, $F, L$ and $R$ represent *"Forward"*, *"Left Turn"* and *"Right Turn"* driving intentions respectively. As shown in Fig. 2, $L_1 = \{F\}, L_2 = \{F, L\}$ and $L_3 = \{F, R\}$.

Amongst all possible interactive situations, we focus on three challenging cases: *Unprotected Left Turn*, *Right Turn Merge*, and *Pedestrian Avoidance*. In these circumstances, the ego vehicle should consider environment information of multiple road entities simultaneously to predict their future behaviors. To avoid accidents and traffic congestion, road geometries, surrounding conditions should also be taken into account. Note that although we consider only vehicles and pedestrians in this problem, our prediction structure is also capable of dealing with other road entities such as cyclists.

We assume that at each time step $t$, we assign an ID, $c_i$, to each car that is within the detection range of our autonomous vehicle. Then, each vehicle's state information can be recorded as $S_{c_i}^t = (x_i^t, y_i^t, \theta_i^t, v_i^t, a_i^t)$, where $x_i^t$ and $y_i^t$ denotes the global position of $c_i$, $\theta_i^t$ denotes the yaw angle, $v_i^t$ and $a_i^t$ denotes the speed and acceleration respectively. Similarly, each pedestrian, identified as $p_i$, has the state information as $S_{p_i}^t = (x_i^t, y_i^t)$. The state of traffic light at time $t$ is denoted by $S_{tl_i}^t \in \{Red, Green, Yellow\}$.

After obtaining the historical information of every dynamically changing object up to the past $T_h$ time steps, our goal is to predict the speed and yaw angle of each vehicle into the future time steps $T_p$. Notice that the future states of each vehicle are highly related to the dynamical environment, which is considered in our proposed prediction structure.

## III. METHODOLOGY

In this section, we first introduce fundamental concepts we used in our method. Then we describe the overall prediction structure we proposed. Finally, feature details of the proposed method are described.

### A. Fundamental Concepts

When we are considering prediction problems, it is important to have probabilistic predictions instead of deterministic results since we need to take into account uncertainties of the road entities. Therefore, the predictor should be based on probabilistic models. In this work, we utilize one of the probabilistic models to demonstrate the effectiveness of our proposed multi-agent interactive prediction (MAIP) system. The model we used is the Conditional Variational Autoencoder (CVAE) [17] [18], which is a latent variable model that is rooted in Bayesian inference and contains an encoder as well as a decoder. The goal of CVAE is to approximate the likelihood distribution:

$$P(Y|X) = \mathcal{N}(\mathcal{F}(z,X), \Sigma_z), \quad (1)$$

where $\mathcal{F}$ is the target function trained by the network to approximate the output $Y$ given some input $X$ and a vector $z$. $\Sigma_z$ denotes the variance of $z$.

During training, input $X$ is fed into the encoder along with a sampled vector from the latent space $z$ which is from some distribution $Q$. According to the CVAE structure, the following equation can be derived:

$$logP(Y|X) - \mathcal{KL}[Q(z|X,Y)||P(z|X,Y)] = \\ E_{z \sim Q}[logP(Y|z,X)] - \mathcal{KL}[Q(z|X,Y)||P(z|X)]. \quad (2)$$

Note that $P(z|X)$ is a prior distribution of latent space, which is always assumed to be normally distributed so that the *KL* divergence can have a closed-form solution. In order to maximize the log likelihood of $P(Y|X)$ on the right-hand-side of Eq. 2, we need to maximize the left-hand-side which becomes the loss function of the CVAE structure:

$$\mathcal{L} = -E_{z \sim Q}[logP(Y|z,X)] + \beta \cdot \mathcal{KL}[Q(z|X,Y)||P(z)] \quad (3)$$

where $Y$ denotes the ground truth, $\hat{Y}$ denotes the output estimation, and $\beta$ is a hyper-parameter that can adjust weights between two loss terms.

### B. Proposed Prediction System

The overall structure of our Multi-agent Interactive Prediction (MAIP) system is shown in Fig. 3. In order to predict future behaviors for every vehicle considering their interactions with the environment, we incorporate input information into two groups: group1 and group2.

Group1 contains information of the entire environment which includes a static map $X_1$ and a dynamic map $X_2$. Group2, on the other hand, considers the information of a single vehicle that we choose to predict. In particular, group2 involves the dynamic map for the selected vehicle, $X_3$, as well as its state information $X_4$, which are extracted from group1. We are then able to predict every entity in the scene by fixing group1 while changing the information in group2 to each corresponding vehicle we want to predict. In this way, any number of road entities can be predicted in parallel under the proposed prediction system. Details of each input will be described in III-C.

To clearly illustrate the framework flow, we take input $X_1$ as a running example. According to Fig. 3, $X_1$ is fed into a convolutional neural network, CNN1, to extract the spatial features of the environment. We define such operation as $f_1(\cdot)$ and then we fed the output into a fully connected layer, FC1, in order to reduce the feature into one dimensional vector $g(W_1 f_1(X_1) + b_1)$, where $g(\cdot)$ denotes the activation function.

Therefore, by considering all four outputs, we are able to obtain the compressed feature information $X_{out}$ as:

$$X_{out} = g\Big(W_5\Big(\sum_{i=1}^{3} g(W_i f_i(X_i) + b_i) \\ + g(W_4 h(X_4) + b_4)\Big) + b_5\Big) \quad (4)$$

where $W_i$ and $b_i$ denote parameters for the network, $h(\cdot)$ denotes mapping function for long short-term memory method. Then $X_{out}$ is fed into CVAE as one of the inputs to the encoder.

Since our goal is to predict future behaviors of each vehicle using its historical observations, each input $\{X_i: i = 2, 3, 4\}$ can expressed as $X_i = X_i^{t-T_h:t}$. Notice that $X_1$ is static map information and will remain the same. Similarly, the corresponding output label for car $c_1$ is denoted as $Y = \tilde{S}_{c_i}^{t+1:t+T_p}$, where $\tilde{S}_{c_i} = (v_{c_i}, \theta_{c_i})$ is a subset of $S_{c_i}$.

By integrating CVAE into our proposed prediction system, the overall loss function of our network can be expressed as:

$$Loss = ||\tilde{S}_{c_i}^{t+1:t+T_p} - \hat{Y}||^2 \\ + \beta \cdot \mathcal{KL}\Big[Q\big(z|X_{out}, \hat{Y}\big) || \mathcal{N}(0,1)\Big]. \quad (5)$$

Moreover, to enable backpropagation, we utilize the reparameterization trick [19] to resolve the non-differential sampling process in the latent space $z$. Note that both the encoder and decoder are used in the training process. However, only the decoder will be used during testing.

### C. Feature Details

In this work, we apply our method in an urban intersection, where interactions and driving conditions are complicated (see Fig. 1). To predict driving behavior of vehicles, road geometries, traffic light information, and the behaviors of heterogeneous surrounding road entities should be taken into account. In our approach, to incorporate all the environment information, we categorize information into four types.

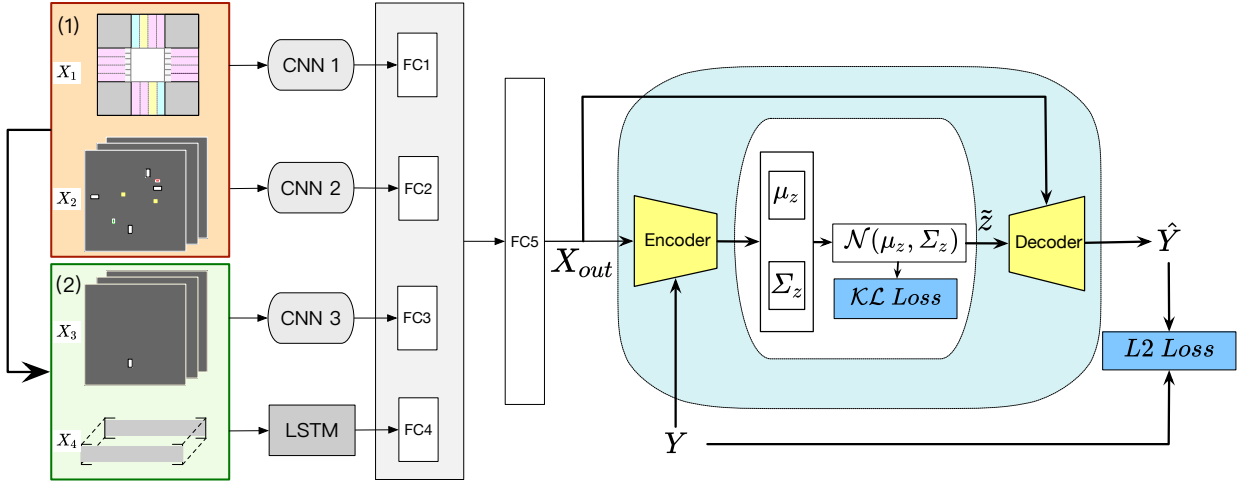

Figure 3. The proposed multi-agent interactive prediction (MAIP) system.

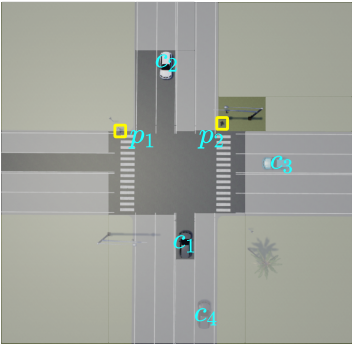

Figure 4. Illustration of the masking mechanism. Pedestrians are bounded by the yellow boxes and the traffic light on the vertical direction is currently green. The unblurred areas represent locations that $c_i$ should pay attention to at the current time step.

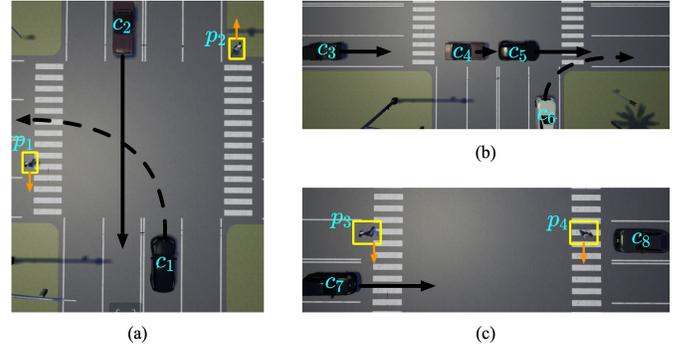

Figure 5. Illustration of three challenging interacting cases under the intersection scenario. Pedestrians are bounded by yellow box and orange arrow represents the moving direction.

1) *Static Map* $(X_1)$: In the static map, we include information that does not change with time such as road geometries, lane information and traffic lights locations. In order to distinguish different types of static information, we assign different numbers in the static grid map which is illustrated by different colors in Fig. 3.
2) *Dynamic Map* $(X_2^t)$: In dynamic map, we consider information that changes with time. To take account interactions with the environment, for a single time step $t$, dynamic information such as the pose of all road entities, and traffic light signals are encoded into the dynamic map. In Fig. 3, vehicles and pedestrians are represented by *white* and *yellow* boxes respectively. Besides, the *green* and *red* boxes denote the traffic light color for the two driving directions.
3) *Dynamic Map for Predicted Vehicle* $(X_3^t)$: focus on dynamic information for the predicted vehicle and information in $(X_3^t)$ are extracted from the dynamic map $(X_2^t)$.
4) *State Information for Predicted Vehicle* $(X_4^t)$: For the predicted vehicle $c_1$ at time $t$, $X_4^t = \{S_{c_i}^t, S_{tl_i}^t\}$ where $S_{c_i}^t$ is the state information of $c_i$, $S_{tl_i}^t$ is the state of traffic light that the predicted vehicle is controlled by.
5) *Mask*: To improve learning efficiency, we introduce the masking mechanism. In fact, each vehicle only needs to focus on nearby environment information that has potential influences on it. Therefore, mask $M^t$ is introduced as

$$\hat{X}_2^t = M^t \odot X_2^t, \qquad (6)$$

where $\odot$ denotes an elementwise product. If the dynamic environment information located in row $i$, column $j$ of $X_2^t$ is taken into account by the ego vehicle, then $M_{ij}^t = 1$, otherwise $M_{ij}^t = 0$. In our approach, we apply the masking mechanism to dynamic map input $X_2^t$. For example, if vehicle $c_1$ is driving towards the intersection area and its front signal light is green, it only focuses on information besides the gray "masked" region as shown in Fig. 4.

## IV. EXPERIMENT

### A. Data Generation

We used the open-source autonomous driving simulator CARLA [20] to simulate a 100 *m* x 100 *m* urban intersection area[1] as shown in Fig. 1 and Fig. 2. Although we used one particular map in this work, for other challenging scenarios, we can either retrain the model or jointly train various scenarios to increase the model's adaptability across different scenes. In the simulation environment, we are able to directly obtain the map information and states of road entities. When applying the model to the real world, we can alternatively obtain the state information processed by Lidar and camera.

Vehicles moving into the intersection are controlled by two traffic light groups whose duration is similar to the real-life situation. Pedestrians are generated with different speeds to pass through each crossroad. To simulate a real-world scenario, pedestrians will keep crossing the road if the traffic light turns red while they are at the center of the road. Therefore, the prediction system is expected to learn that pedestrians should have higher priority than the traffic light for each on-road vehicle.

To record data in various scenarios, we randomly generate four to six cars in the map and their driving behaviors are automatically controlled by the simulator. We collected 37, 265 frames for training in total with sample frequency of 5*HZ*. Among all the interacting situations, we focus on three challenging cases (Fig. 5):

1) *Unprotected Left Turn:* An unprotected left turn occurs at an intersection where there is no traffic light to signal the left turn (see Fig. 5 (a)). The critical rule for the vehicle to make a left turn is that its driving behaviors do not influence other vehicles moving forward from the opposite side of the road. In other words, straight driving vehicles have higher road priority than vehicles to make a left turn in intersection areas to avoid accidents. Hence, car $c_2$ will not make a left turn until car $c_1$ passes through the intersection.

2) *Right Turn Merge:* As it is shown in Fig. 5 (b), vehicles driving in two directions merge into the same lane. In our scenarios, there is no traffic signal for the right turn and vehicles are able to turn if it is safe. Generally, vehicles with right turn intention driving have low road priority and they should wait until it is safe to turn. In this example, $c_6$ is more likely to insert in front of $c_3$ than of $c_5$ since the gap between $c_3$ and $c_4$ is larger.

3) *Pedestrian Avoidance:* Since we want to consider the interaction between vehicles and pedestrians, we place the pedestrians on two crosswalks in the scene. When vehicles are driving towards pedestrians, they are expected to brake ahead of the time to avoid accidents. For the case shown in Fig. 5 (c), $c_8$ should stop since it has a strong interaction with $p_4$. However, for $c_7$, its behavior is less influenced by $p_3$ and it can choose to either pass or yield the pedestrian.

### B. Implementation Details

In this section, we introduce the implementation details of our MAIP system. For all the three convolutional neural networks, we utilize a 3 x 3 kernel. The long-short-term memory (LSTM) network of 16 hidden neurons is utilized to incorporate historical state information with time for the input $X_4$. Also, we apply fully connected layers FC1 to FC4 of 8 neurons each, and FC5 of 16 neurons. For the CVAE structure, we use two fully connected layers of 16 neurons for our encoder and decoder. According to the cross-validation results, we utilize a two dimensional latent z space to map our high dimensional input features into a low dimensional space.

### C. Evaluation

*1) Quantitative Results:* To evaluate the performance of our approach, Root Mean Square Error (RMSE) is adapted as the evaluation metric for our system. We compare our method with other four methods as follows:

- *IDM (Intelligent Driver Model).* This method is a simple car following approach where the ego vehicle only considers interaction with its front vehicle.
- *CNN-LSTM.* In this method, future behaviors of vehicles are predicted by using similar structure from our approach without CVAE. While future behaviors predicted by this method are deterministic, in order to calculate its standard deviation and compare the results with other probabilistic models, the ensemble method was utilized.
- *CNN-CVAE.* This method is based on CVAE and it has a similar framework structure as our approach except that the state information for ego vehicle $X_4$ is fed into fully connected layers instead of LSTM.
- *MAIP-Recursive.* This method predicts future behaviors $S_i^{t+1:t+T_p}$ for the ego vehicle by setting $T_p$ to a single frame. Then we feed the predicted behavior to the network to predict $S_i^{t+1:t+T_p+1}$ recursively.

To analyze performance amongst different methods at each time step *t*, we compare *RMSE* and its standard deviation for both yaw angle $\theta$ and speed *v* under different prediction time steps. Results are shown in Table I. The standard deviation for *IDM* is not shown because it is a deterministic model which cannot make probabilistic predictions. For all the methods, prediction errors are accumulated with time step *t*. It is clear that the prediction performance of *IDM* is far from others. The reason is that *IDM* is a simple car following model which means the predicted vehicle only considers the interaction with the vehicle ahead. Thus, this method is very limited when the predicted vehicle has interactions with multiple vehicles and traffic lights.

---
[1] The simulation environment and some testing results can be found on https://youtu.be/aWMR9I8yuhg

TABLE I. Evaluation for different methods. Speed (v) unit is m/s and yaw angle (θ) is measured in degree.

| Method | | 0.2s | 0.4s | 0.6s | 0.8s | 1s |
|---|---|---|---|---|---|---|
| IDM | $\theta$ | 24.73 | 34.72 | 42.20 | 48.65 | 54.26 |
|  | $v$ | 0.49 | 0.63 | 0.83 | 0.99 | 1.12 |
| CNN-LSTM | $\theta$ | 6.82 ± 0.16 | 7.29 ± 0.19 | 7.65 ± 0.14 | 8.43 ± 0.15 | 9.32 ± 0.26 |
|  | $v$ | 0.47 ± 0.03 | 0.71 ± 0.02 | 0.78 ± 0.03 | 0.95 ± 0.03 | 1.06 ± 0.04 |
| CNN-CVAE | $\theta$ | 7.99 ± 0.51 | 8.29 ± 0.82 | 7.55 ± 0.86 | 8.80 ± 0.95 | 9.67 ± 1.22 |
|  | $v$ | 0.95 ± 0.11 | 1.01 ± 0.12 | 0.98 ± 0.13 | 1.03 ± 0.17 | 1.15 ± 0.20 |
| MAIP-Recursive | $\theta$ | **4.91 ± 0.68** | 6.32 ± 0.62 | 8.36 ± 0.11 | 9.54 ± 0.13 | 11.41 ± 2.38 |
|  | $v$ | **0.41 ± 0.05** | 0.77 ± 0.09 | 0.85 ± 0.08 | 0.91 ± 0.11 | 1.03 ± 0.12 |
| MAIP (our method) | $\theta$ | 5.20 ± 0.70 | **5.23 ± 0.73** | **5.38 ± 0.81** | **6.06 ± 0.85** | **7.19 ± 1.03** |
|  | $v$ | 0.56 ± 0.04 | **0.61 ± 0.05** | **0.63 ± 0.05** | **0.68 ± 0.06** | **0.82 ± 0.08** |

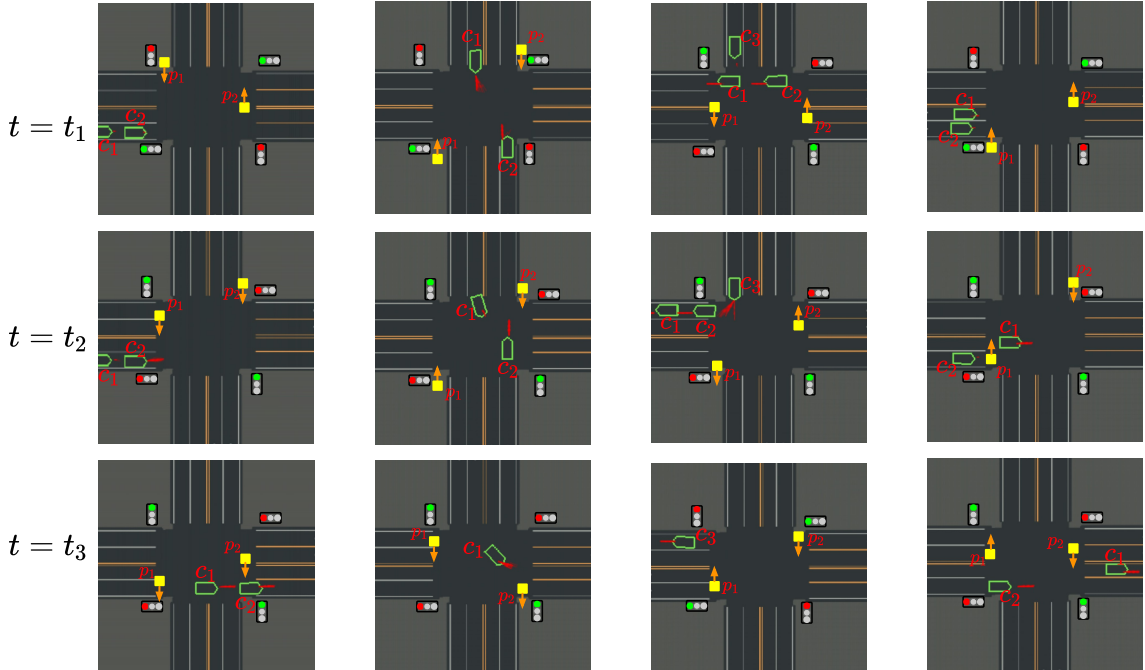

Figure 6. Illustration of results applying MAIP approach in different driving scenarios. Vehicles and pedestrians are represented by *green* and *yellow* bounding boxes respectively. The orange arrow on each pedestrian represents the walking direction. Predicted trajectories for each vehicle are represented by red lines. For clear visualization, at each time step, twenty predicted trajectories are sampled from our MAIP prediction system for each vehicle.

Moreover, since the geometry information for scenarios is not taken into account by *IDM*, RMSE for steering angle is nearly eight times bigger than *MAIP*.

Comparing the results between *CNN-CVAE* and *MAIP*, the former approach is apparently less accurate and has a higher standard deviation, which indicates that *LSTM* can better incorporate historical information in future behavior prediction problems especially when the predicted horizon gets longer. Comparing the results between *CNN-LSTM* and *MAIP*, the latter approach is more accurate and has smaller standard deviation, which shows that the direct ensemble method doesn't achieve a better performance than the learning-based probabilistic model.

It should be noted that the standard deviation of *CNN-LSTM* is smaller than other models using CVAE because *CNN-LSTM* method does not take into account multimodal distribution. For a short time step $t$, *MAIP-Recursive* is more accurate than *MAIP*, but for a longer prediction sequence, its errors accumulate and lead to inaccurate prediction.

*2) Visualization Results:* In order to verify that our prediction system has the ability to incorporate environmental information into various scenarios, we choose four representative test cases that are not seen in the training data for illustration, which is shown in Fig. 6.

- **Case A:** In this case, it can be verified that our system is able to incorporate traffic light information and interactions between vehicles driving in the same direction. At time step $t_1$, car $c_1$ and $c_2$ are predicted to stop because of the Red traffic light. As the traffic signal changes to Green at $t_2$, $c_1$ is predicted to move forward while $c_2$ is predicted to follow c1 with a smaller velocity until their distance becomes larger at $t_3$.

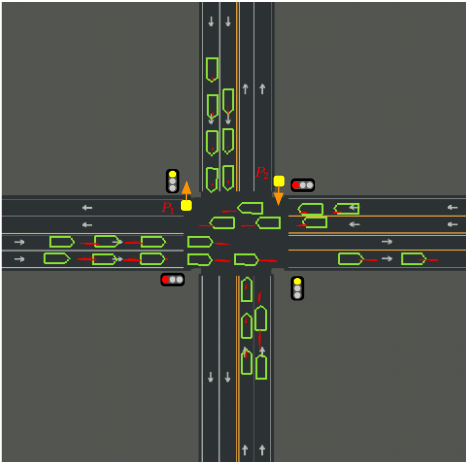

Table 7. Illustration of results applying MAIP approach in a complex driving scenario with high traffic density.

- **Case B:** This case illustrates the *Unprotected Left Turn* scenario and indicates that our approach can take account of road geometry and interaction among vehicles driving in different directions. As can be seen through the second column in Fig. 6, $c_1$ is predicted to not make a left turn until $c_2$ passes through the center area of the intersection. At the same time, $c_2$ is predicted to remain its velocity without yielding $c_1$ since our system learned that $c_2$ has the right-of-way.
- **Case C:** In this case, the ability of our prediction system under dense traffic in a *Right Turn Merge* scenario is examined. As shown in the third column of Fig. 6, the gap between $c_1$ and $c_2$ is not big enough. Thus, $c_3$ is predicted to merge until both cars on the main road pass.
- **Case D:** Here we examine the proposed system's prediction ability when strong interactions between vehicles and pedestrians occur. According to Fig. 6, at time step $t_1$, a pedestrian $p_1$ is about to cross the road and its location is closer to $c_2$ than to $c_1$. Therefore, at $t_2$, $c_2$ is predicted to break in front of $p_1$ while c1 is predicted to keep driving forward due to the green traffic light.

Moreover, we also test the robustness of our MAIP system under a more complicated driving scenario with 29 vehicles. As can be seen in Fig. 7, the MAIP model is able to predict any number of road entities and take interactions into account even when the system is trained on a much smaller number of entities.

According to the selected cases, our approach is capable of predicting future behaviors for vehicles driving in various situations considering the interactions with other road entities. Moreover, for vehicles driving in $L_2$ and $L_3$ under *Case B* and *Case C*, they can either go straight or make a turn. Therefore, it is expected that when vehicles drive closer to the intersection center, the predictor should consider prediction uncertainties by considering every possibility of the driving directions. According to the results, the sampled trajectories indeed split into different groups which illustrate that the desired multimodal distribution can be predicted by our proposed framework.

## V. CONCLUSIONS

In this paper, a Multi-agent Interactive Prediction (MAIP) system is proposed, which utilizes static and dynamic environment information to predict every road entity while taking into account their mutual interaction. We examined the performance of our proposed system under a simulated urban intersection scenario. We first compared the prediction accuracy of our method with four different approaches and concluded that the proposed MAIP system outperforms others in terms of the mean and standard deviation of the prediction error especially for long prediction horizons. We then selected four representative testing cases to illustrate the capability of our method under various unseen challenging scenarios. The result shows that the proposed MAIP system successfully learned to reason about complicated environment information and provide rational prediction results for every on-road vehicle. In future work, we will extend the current allo-centric input to the egocentric input, which may provide more robustness and adaptivity to our system.